\documentclass[conference]{IEEEtran}
\IEEEoverridecommandlockouts
\usepackage{cite}
\usepackage{amsmath,amssymb,amsfonts}
\usepackage{algorithmicx}

\usepackage{graphicx}
\newtheorem{definition}{Definition}[section]
\usepackage{xfrac}

\usepackage{bm}
\usepackage{textcomp}
\usepackage{xcolor}
\usepackage{comment}
\usepackage{algorithm}
\usepackage{algpseudocode}
\usepackage{enumitem}

\usepackage[a4paper, total={184mm,239mm}]{geometry}
\def\BibTeX{{\rm B\kern-.05em{\sc i\kern-.025em b}\kern-.08em
    T\kern-.1667em\lower.7ex\hbox{E}\kern-.125emX}}

\makeatletter
\newcommand{\newlineauthors}{%
  \end{@IEEEauthorhalign}\hfill\mbox{}\par
  \mbox{}\hfill\begin{@IEEEauthorhalign}
}
\makeatother

\begin{document}

\title{BOiLS: Bayesian Optimisation for Logic Synthesis}

\author{\IEEEauthorblockN{Antoine Grosnit*\thanks{* Equal contribution.}}
\IEEEauthorblockA{\textit{Huawei Noah's Ark Lab} \\
antoine.grosnit@huawei.com}
\and
\IEEEauthorblockN{Cedric Malherbe*}
\IEEEauthorblockA{\textit{Huawei Noah's Ark Lab} \\
cedric.malherbe@huawei.com}
\and
\IEEEauthorblockN{Rasul Tutunov}
\IEEEauthorblockA{\textit{Huawei Noah's Ark Lab} \\
rasul.tutunov@huawei.com}
\newlineauthors
\IEEEauthorblockN{Xingchen Wan}
\IEEEauthorblockA{\textit{Huawei Noah's Ark Lab} \\
xingchen.wan@huawei.com}
\and
\IEEEauthorblockN{Jun Wang}
\IEEEauthorblockA{\textit{Huawei Noah's Ark Lab} \\
University College London \\
w.j@huawei.com}
\and
\IEEEauthorblockN{Haitham Bou Ammar}
\IEEEauthorblockA{\textit{Huawei Noah's Ark Lab} \\
University College London \\
haitham.ammar@huawei.com}
}

\maketitle

\begin{abstract}
Optimising the quality-of-results (QoR) of circuits during logic synthesis is a formidable challenge necessitating the exploration of exponentially sized search spaces. While expert-designed operations aid in uncovering effective sequences, the increase in complexity of logic circuits favours automated procedures. Inspired by the successes of machine learning, researchers adapted deep learning and reinforcement learning to logic synthesis applications. However successful, those techniques suffer from high sample complexities preventing widespread adoption. 

To enable efficient and scalable solutions, we propose BOiLS, the first algorithm adapting modern Bayesian optimisation to navigate the space of synthesis operations. BOiLS requires no human intervention and effectively trades-off exploration versus exploitation through novel Gaussian process kernels and trust-region constrained acquisitions. In a set of experiments on EPFL benchmarks, we demonstrate BOiLS's superior performance compared to state-of-the-art in terms of both sample efficiency and QoR values. 
\end{abstract}


\section{Introduction}
During the pre-mapping stages of logic synthesis, designers uncover a series of structural transformations that improve circuit efficiencies by maximising performance criteria, such as the Quality-of-Results (QoR)~\cite{QoR_max,  review_logic_synthesis}. Modernistic synthesis tools administer those transformations by first representing circuits as And-Inverter Graphs (AIGs) and then employing technology-independent operations to reduce graph sizes while adhering to delay constraints. Although experts devised a plethora of QoR optimisers~\cite{himap, pimap}, exponentially sized exploration spaces, especially in large circuits, still pose formidable challenges to the design of predefined synthesis flows. The quest for scalable and sample efficient solvers has, in turn, stimulated novel research trends that benefit from state-of-the-art developments in machine learning (ML) when tailored to logic synthesis applications. 
 
Although ML techniques emerged as active areas of research within the holistic electronic design automation pipeline (e.g., in design space reduction~\cite{ml_design_reduction}, placement~\cite{ml_placement}, routing~\cite{ml_routing}, testing and verification~\cite{ml_testing, ml_verification} and in manufacturing),  their examination in logic synthesis only recently started to gain attention. Ere to this work, the authors in~\cite{survey_ml_for_eda} distinguish a handful of ML-inspired approaches based on deep neural networks and reinforcement learning to obtain optimal structural transformations. For instance, the work in~\cite{cnn_for_eda} adopts (deep) convolutional neural networks to solve multi-class classification problems mapping synthesis flows to QoR levels. Recently~\cite{yu2020decision} also propose an LSTM-based approach for QoR optimisation. The authors in~\cite{drills, GCN_for_eda}, on the other hand, extend deep reinforcement learning (DRL) to pre-mapping applications by defining novel Markov decision processes and policies that capture the intricate complexities of logic synthesis. 

\begin{figure}[t]
\label{fig:sample_efficiency}
\centering
  \includegraphics[trim=2em 3em 3em 2em, width=0.4\textwidth]{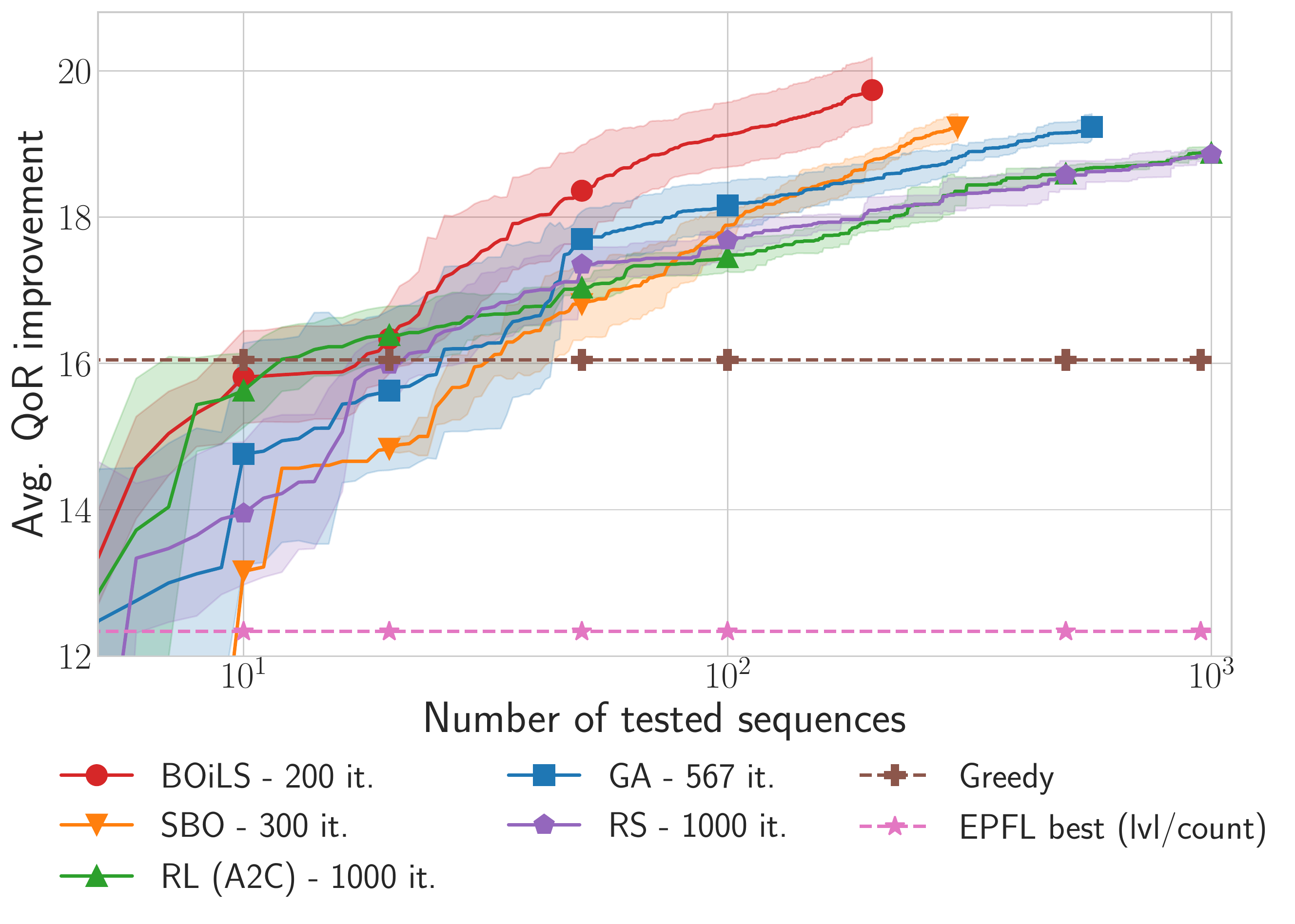}
  \caption{Average QoR results over 10 EPFL circuits to recover $97.5\%$ of the QoR values achieved by BOiLS after only \emph{200} sequence trials. We notice that BOiLS attains best QoR values while requiring $1.5$ fewer evaluations than standard Bayesian optimisation (SBO), $2.8$ times less compared to genetic algorithms (GA), and over $5$ times with respect to deep reinforcement learning.}
  \vspace{-6mm}
\end{figure}
Albeit their widespread usage, both deep learning and reinforcement learning techniques exhibit \emph{high sample (data) complexities}~\cite{model_based_atari} especially in high-dimensional combinatorial spaces. When applied to logic synthesis, such high data demands amount to numerous evaluations within a given circuit, e.g., 10,000 sequences per circuit when adopting convolutional deep networks~\cite{cnn_for_eda}, or over thousands of agent environment interactions in DRL (see Section~\ref{sec:experiments}).   

\underline{\textbf{Contributions:}} This paper contributes to the above problems by introducing BOiLS, the first Bayesian optimisation (BO) solver for logic synthesis. BOiLS demands no human intervention and efficiently searches combinatorial spaces by trading-off exploration versus exploitation. On a high level, our method operates in two steps. First, we fit a surrogate Gaussian process (GP) to QoR data utilising kernels geared towards transformation sequences of AIG graphs. This GP enables both sample efficiency and calibrated uncertainty estimation, which we then exploit in the second step to suggest new synthesis flows for evaluation. Here, we harness concepts from local trust-region acquisition function maximisation to effectively handle high-dimensionalities. In a set of experiments on the comprehensive EPFL benchmark~\cite{epfl_benchmark}, we demonstrate superior QoR performance and better sample efficiencies compared to deep reinforcement learning~\cite{drills}, graph neural network policies~\cite{GCN_for_eda}, genetic algorithms and other search strategies, as well as against the best results from the EPFL leadership board~\cite{epfl_benchmark} in 8 out of 10 circuits. Succinctly, our contributions can be summarised as follows: \textit{i)} Formulating logic synthesis as an instance of black-box combinatorial optimisation; \textit{ii)} Proposing BOiLS as the first Bayesian optimisation solver for logic synthesis applications; \textit{iii)} Proposing AIG tailored Gaussian process kernels and acquisition optimisers that handle high-dimensional search spaces; \textit{iv)} Outperforming state-of-the-art techniques in 8 out of 10 circuits from the EPFL benchmark~\cite{epfl_benchmark}; and \textit{v)} Open-sourcing code to ease the reproducibility of our findings in case of acceptance.

\section{Problem Definition}
\label{sec:problem_statement}

In logic synthesis we aim to find an equivalent yet simpler representation of a logic design using a series of primitive transformations. Modern tools~\cite{ABC, yosys} express a circuit, $\mathcal{C}$, as a directed and acyclic graph, referred to as an AIG, to denote a structural implementation of the circuit's logical functionality. AIGs consist of two-input nodes representing logical conjunction, terminal nodes labelled with variable names, and edges (optionally) containing markers indicating logical negation~\cite{LazyMan}. Our goal is to uncover a sequence $\texttt{seq} = [\texttt{s}_1, \dots, \texttt{s}_K] \in \texttt{Alg}^{K}$ of at most $K>1$ operations to optimise the graph's structure. Here, $\texttt{Alg}=\{\texttt{A}_1, \dots, \texttt{A}_{n}\}$ denotes a set of $n$ transformation algorithms (e.g., \texttt{resub}, \texttt{rewrite}, \texttt{refactor}; see~\cite{ABC} for the full list of possible operations) that can be executed to alter the AIG. 

We employ QoR to assess the performance of an evaluated sequence. Precisely, having applied $\texttt{seq}$ to an AIG, we register both the area, $\text{Area}_{\mathcal{C}}(\texttt{seq})$, and the delay, $\text{Delay}_{\mathcal{C}}(\texttt{seq})$ after executing FPGA mapping. Specifically, we correspond $\text{Area}_{\mathcal{C}}(\texttt{seq})$ to the number of lookup tables used for mapping the AIG (LUT-count) and $\text{Delay}_{\mathcal{C}}(\texttt{seq})$ to the longest path between primary inputs and outputs of the resulting graph (Levels). Then, we compute the overall effectiveness of $\texttt{seq}$:
\begin{equation}
\label{Eq:QoR}
\text{QoR}_\mathcal{C}(\texttt{seq}) = \frac{\text{Area}_\mathcal{C}(\texttt{seq})}{\text{Area}_\mathcal{C}(\texttt{ref})} + \frac{\text{Delay}_\mathcal{C}(\texttt{seq})}{\text{Delay}_\mathcal{C}(\texttt{ref})},
\end{equation}
where $\text{Area}_\mathcal{C}(\texttt{ref})$ and $\text{Delay}_\mathcal{C}
(\texttt{ref})$ denote area and delay of a resulting application of a reference sequence (e.g., $\texttt{resyn2}$~\cite{ABC}). Our QoR definition in Equation~(\ref{Eq:QoR}) is reasonably standard, measuring relative decrements in area and delay against reference series of operations. Hence, an optimal  $\texttt{seq}^{\star}$ is that sequence in $\texttt{Alg}^{K}$ which minimises $\text{QoR}_{\mathcal{C}}(\texttt{seq})$: 
\begin{equation}
\label{eq:best_solution}
    \texttt{seq}^{\star} \in 
    \underset{\texttt{seq} \in \texttt{Alg}^{K}}{\arg\min}~\text{QoR}_{\mathcal{C}}(\texttt{seq}) \equiv \underbrace{{\arg\max}~\text{-QoR}_{\mathcal{C}}(\texttt{seq})}_{\text{This paper's focus}}.
\end{equation}
\underline{\textbf{Finding $\texttt{seq}^{\star}$:}} We note two difficulties when seeking $\texttt{seq}^{\star}$: 
\begin{itemize}[leftmargin=0.15in, noitemsep, topsep=0.05pt]
    \item \textbf{QoRs as Black-Box Functions:} From Equation~(\ref{Eq:QoR}), we discern that a closed analytical form of $\text{QoR}_{\mathcal{C}}(\cdot)$ as a function of $\texttt{seq}$ is challenging to obtain for any circuit $\mathcal{C}$. Such difficulties stem from the fact that QoR computations involve complex algorithmic processes executed over AIG graphs. For example, operation applications, as well as area and delay calculations perform highly optimised C/C++ instructions; see~\cite{ABC}. 
    In machine learning, we refer to \emph{functions of unknown analytical} forms as \emph{black-boxes} and seek \emph{efficient data-driven solvers} that optimise for $\texttt{seq}^{\star}$ based on a handful of sequence evaluations. 
    \item \textbf{Exponentially Sized Search Spaces:} In attempting a data-driven solution, we remark an exponential growth in the search space even for one circuit $\mathcal{C}$. Specifically, since the cardinality of the search space $|\texttt{Alg}^K|$ equates to $n^K$ ($11^{20}$ in our experiments), an exhaustive exploration of $\texttt{Alg}^K$ is unrealistic in practice. Moreover, due to the black-box nature of $\text{QoR}(\cdot)$, it is difficult to assume desirable characteristics like linearity or submodularity that facilitate searching for $\texttt{seq}^{\star}$~\cite{submodular_opt}. Hence, the problem in Equation~(\ref{eq:best_solution}) is generically combinatorial 
    by nature making it impossible to design exact solvers without exploring the whole search space.
\end{itemize}

\section{Bayesian Optimisation for Logic Synthesis}
\label{sec:bayesian_opt}

\subsection{Primer on Bayesian Optimisation \& Gaussian Processes}\label{Sec:Primal}
BO is a gradient-free technique used to optimise expensive-to-evaluate black-box functions. BO tackles global optimisation sequentially, where at each round $t$, the learner selects an input probe $\bm{x}_t$ for evaluation and acquires a corresponding (noisy) black-box function value $g(\bm{x}_t)$. Typically, inputs and outputs take on continuous values in bounded domains, whereby $\bm{x}_t \in \mathcal{X} \subseteq \mathbb{R}^{d}$ with $d$ denoting the search space's dimensionality and $g(\bm{x}_t) \in \mathbb{R}$. The goal is to \emph{rapidly (in terms of function evaluations)} approach the maximum $\bm{x}^{\star} = \arg\max_{\bm{x}\in\mathcal{X}} g(\bm{x})$~\cite{review_bayes_opt, HEBO, casmo}. To achieve the above goal, BO relies on historical data (e.g., $\left\langle \bm{x}_1, g(\bm{x}_1) \right\rangle, \dots, \left\langle \bm{x}_{t}, g(\bm{x}_{t}) \right\rangle$ at round $t$) to \emph{i)} build a surrogate of the actual black-box and \emph{ii)} utilise the learnt surrogate to decide on the new input probe to evaluate in the subsequent round. Since both $g(\cdot)$ and $\bm{x}^{\star}$ are unknown, learners need to trade off exploitation and exploration during the search process. A natural way of handling this dilemma is basing decisions on the surrogate's \emph{predictive distribution}, where we contrast fully trusting the surrogate's mean prediction or examining unseen inputs. Formalising such choices in BO is accomplished via maximising acquisition functions that we survey in Section~\ref{Sec:Acq}. Equipped with a \emph{probabilistic model} and an \emph{acquisition function}, a generic BO template of the above steps is shown in Algorithm \ref{alg:BO}. 

\begin{algorithm}[ht]
\begin{algorithmic}[1]
\State \textbf{Inputs:} Budget, initial data set $\mathcal{D}_0= \{\bm{x}_l, y_l\equiv g(\bm{x}_l)\}_{l=1}^{n_0}$
\For{$t=0, \dots, \text{Budget}-1$}
\State Use data to fit a surrogate probabilistic model 
\State Determine $\bm{x}_\text{new}$ by maximising an acquisition function 
\State Evaluate new probes acquiring $y_{\text{new}}\equiv g(\bm{x}_\text{new})$
\State Augment data $\mathcal{D}_{t+1} = \mathcal{D}_{t} \cup  \left\langle\bm{x}_\text{new}, y_{\text{new}}\right\rangle$
\EndFor 
\State \textbf{Output:} $\bm{x}^{\star} \in \arg\max_{\bm{x} \in \mathcal{D}_{\text{Budget}}} g(\bm{x})$.  
\end{algorithmic}
\caption{Template of Bayesian Optimisation Algorithms}
\label{alg:BO}
\end{algorithm}

\subsubsection{Probabilistic Modelling \& Gaussian Processes} As designated in line 3 of Algorithm 1, the first step involves fitting a surrogate model that provides well-calibrated uncertainty estimates and is efficient in terms of black-box evaluations. Among various machine learning candidates, Gaussian processes (GPs), offer a flexible and sample-efficient procedure for placing priors over unknown functions~\cite{Rasmussen}. Formally, a GP is defined as: 
\begin{definition}[Gaussian Process~\cite{Rasmussen}]\label{Def:GP}
A GP is an \emph{infinite collection of random variables any finite number of which have a joint Gaussian distribution}.
\end{definition}

We can use GPs to directly define distributions over functions, where we write $g(\bm{x}) \sim \mathcal{G}\mathcal{P}(m(\bm{x}), k(\bm{x}, \bm{x}^{\prime}))$. Here, $m(\bm{x}) = \mathbb{E}[g(\bm{x})]$ and $k(\bm{x}, \bm{x}^{\prime}) = \mathbb{E}[(g(\bm{x})-m(\bm{x}))(g(\bm{x}^{\prime})-m(\bm{x}^{\prime}))]$ denote  the mean and covariance functions that fully specify a GP. Following~\cite{Rasmussen}, we set the mean function to zero, thus having $g(\bm{x}) \sim \mathcal{G}\mathcal{P}(0, k(\bm{x}, \bm{x}^{\prime}))$. 

\textbf{Covariance kernels} encode our (smoothness) assumptions about the function $g(\bm{x})$ that we wish to learn. GP kernels usually impose a similarity postulate that close input points are likely to have similar target values. That is, GPs measure the covariance between $g(\bm{x})$ and $g(\bm{x}^{\prime})$ as a decreasing function of the distance between the two inputs $\bm{x}$ and $\bm{x}^{\prime}$, i.e., $\operatorname{Cov}(g(\bm{x}), g(\bm{x}^{\prime})) 
\equiv
k(\bm{x}, \bm{x}^{\prime}) = \Psi(\operatorname{d}(\bm{x}, \bm{x'}))$ for some decreasing function $\Psi$ and distance function $\operatorname{d}(\cdot,\cdot)$. 
In terms of the kernel choice, there are a wide array of options with squared exponential (SE) $k_{\text{SE}}(\cdot, \cdot)$, and Mat\'ern(5/2) being the most common in BO~\cite{HEBO}. 
Throughout our exposition, we focus on $k_{\text{SE}}(\bm{x}, \bm{x}^{\prime})$ that measures covariances as a function of L2 distances between two inputs such that the closer $\bm{x}$ gets to $\bm{x}^{\prime}$, the higher the correlation between $g(\bm{x})$ and $g(\bm{x}^{\prime})$, i.e., $k_{\text{SE}}(\bm{x}, \bm{x}^{\prime}) \propto \exp\left(-\sfrac{||\bm{x} - \bm{x}^{\prime}||_{2}^{2}}{2}\right)$. 
Given a finite set of input data points $\bm{x}_{1:n}\equiv \{\bm{x}_{i}\}_{i=1}^{n}$, we can now utilise Definition 3.1 to derive the jointly Gaussian prior distribution on the corresponding outputs $\bm{g} \equiv \{g(\bm{x}_{i})\}_{i=1}^{n}$: $
    \bm{g} \sim \mathcal{N}\left(\bm{0}, \bm{K}(\bm{x}_{1:n}, \bm{x}_{1:n})\right)$, where $\bm{K}(\bm{x}_{1:n}, \bm{x}_{1:n}) \in \mathbb{R}^{n\times n}$ is the covariance matrix with its $(i,j)^{th}$ entry computed as $[\bm{K}(\bm{x}_{1:n}, \bm{x}_{1:n})]_{i,j} = k(\bm{x}_{i}, \bm{x}_j)$. 

\textbf{Predictions using GPs:} 
Given training input-output observations $\{\bm{x}_i, g(\bm{x}_i)\}_{i=1}^{n}$, we would like to construct the \emph{output predictive distributions} at $\tilde{n}$ test points $\{\bm{\tilde{x}}_{j}\}_{j=1}^{\tilde{n}}$. Assuming that training and test outputs share the same data generating distribution, as is the case in any supervised learning setting, the joint distribution over training and testing function values $\bm{g}$ and $\bm{\tilde{g}} \equiv \{g(\bm{\tilde{x}}_{j})\}_{j=1}^{\tilde{n}}$ follows: 
\begin{equation*}
    \left[\begin{array}{c}
         \bm{g}\\
         \bm{\tilde{g}}
    \end{array}\right] \sim \mathcal{N}\left(\bm{0}, \left[\begin{array}{cc}
             \bm{K}(\bm{x}_{1:n}, \bm{x}_{1:n}) & \bm{K}(\bm{x}_{1:n}, \bm{\tilde{x}}_{1:\tilde{n}}) \\
              \bm{K}^{\mathsf{T}}(\bm{x}_{1:n}, \bm{\tilde{x}}_{1:\tilde{n}}) & \bm{K}(\bm{\tilde{x}}_{1:\tilde{n}}, \bm{\tilde{x}}_{1:\tilde{n}}) 
         \end{array}\right]\right),
\end{equation*}
where $\bm{K}(\bm{x}_{1:n}, \bm{\tilde{x}}_{1:\tilde{n}}) \in \mathbb{R}^{n\times \tilde{n}}$ and $\bm{K}(\bm{\tilde{x}}_{1:\tilde{n}}, \bm{\tilde{x}}_{1:\tilde{n}}) \in \mathbb{R}^{\tilde{n}\times \tilde{n}}$ denote the covariances matrices evaluated at all pairs of training and test points and those between test points. To arrive at predictive output distributions, we condition the above multi-variate Gaussian leading to:
\begin{equation}
\label{Eq:PredictiveDistribution}
    \bm{\tilde{g}}|\bm{g}, \{\bm{x}_{i}\}_{i=1}^{n}, \{\bm{\tilde{x}}_{j}\}_{j=1}^{\tilde{n}} \sim \mathcal{N}\left(\bm{\mu}_{\text{posterior}}, \bm{\Sigma}_{\text{posterior}}\right),
\end{equation}
where the posterior mean and covariance are given by: 
\begin{align*}
    \bm{\mu}_{\text{posterior}} & = \bm{K}(\bm{x}_{1:\tilde{n}}, \bm{x}_{1:n})\bm{K}^{-1}(\bm{x}_{1:n}, \bm{x}_{1:n})\bm{g} \\ 
    \bm{\Sigma}_{\text{posterior}} & =\bm{K}(\bm{x}_{1:\tilde{n}}, \bm{x}_{1:\tilde{n}}) \\
    & \hspace{1em}-  \bm{K}(\bm{x}_{1:\tilde{n}}, \bm{x}_{1:n})\bm{K}^{-1}(\bm{x}_{1:n}, \bm{x}_{1:n})\bm{K}(\bm{x}_{1:n}, \bm{x}_{1:\tilde{n}}).
\end{align*}

\textbf{Learning in GPs:} So far, we have specified a probabilistic framework capable of producing output predictions on unseen inputs. The remaining ingredient in a GP pipeline involves introducing the kernel hyperparameters that are tuned using marginals to fit a given dataset best. In SE kernels, for example, we can inject a length-scale parameter per each input-dimension writing: $k_{\bm{\theta}}^{\text{SE}}(\bm{x}, \bm{x}^{\prime}) = \exp (-\sfrac{1}{2}r^{2})$ with $r= \sqrt{(\bm{x} - \bm{x}^{\prime})^{\mathsf{T}}\text{diag}(\bm{\theta}^{2})^{-1}(\bm{x} - \bm{x}^{\prime})}$. Here, $\bm{\theta} \in \mathbb{R}^{d}$ denotes the $d$ length-scale hyperparameters that need to be fit, such that $\bm{\theta}^{2}$ is executed element-wise and $\text{diag}(\bm{v})$ represents a diagonal matrix of a vector $\bm{v}$. In standard GPs~\cite{Rasmussen}, $\bm{\theta}$ are determined by minimising the negative log marginal likelihood leading us to the following optimisation problem: 
\begin{equation}
\label{Eq:NLLM}
    \min_{\bm{\theta}} \mathcal{J}(\bm{\theta}) = \frac{1}{2} \text{det}(\bm{K}_{\bm{\theta}}(\bm{x}_{1:n}, \bm{x}_{1:n})) + \frac{1}{2} \bm{g}^{\mathsf{T}}\bm{K}^{-1}_{\bm{\theta}}(\bm{x}_{1:n}, \bm{x}_{1:n})\bm{g},
\end{equation}
where $\text{det}(\bm{K}_{\bm{\theta}}(\bm{x}_{1:n}, \bm{x}_{1:n}))$ is the determinant of the covariance matrix $\bm{K}_{\bm{\theta}}(\bm{x}_{1:n}, \bm{x}_{1:n})$ such that $[\bm{K}_{\bm{\theta}}(\bm{x}_{1:n}, \bm{x}_{1:n})]_{i,j} = k_{\bm{\theta}}^{\text{SE}}(\bm{x}_i, \bm{x}_{j})$. To remedy the need to invert an $n\times n$ covariance matrix, one can follows new advancements in modern GPs~\cite{James}. 
\begin{figure}[ht]
    \centering
    \includegraphics[trim=4em 3em 4em 3em, width=.6\columnwidth]{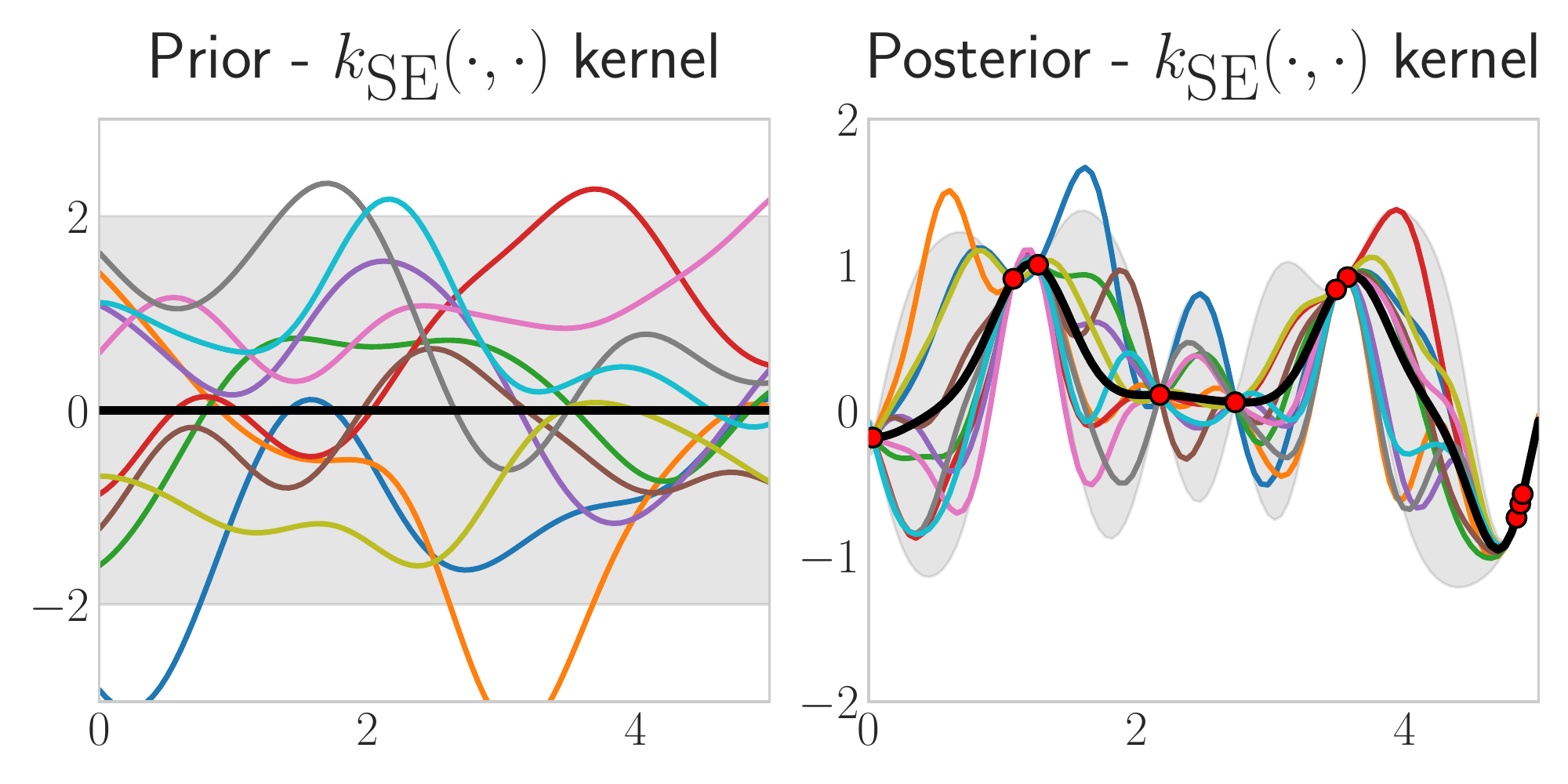}
    \caption{(Left) Samples generated from a GP priors with $k_{\text{SE}}(\cdot)$ before observing any data. (Right) Samples from the GP posterior (Equation~(\ref{Eq:PredictiveDistribution})) after training the kernel's hyperparameters (Equation~(\ref{Eq:NLLM})).}
    \label{fig:prop}
\end{figure}

\subsubsection{Acquisition Functions}\label{Sec:Acq}
Having introduced a distribution over latent black-box functions, we now discuss the process by which novel query points are
suggested for collection in order to improve the surrogate model’s best guess for the global $\bm{x}^{\star}$. In BO, proposing novel query points is performed through maximising an acquisition function that trades off exploration and exploitation using the fitted GP's posterior distribution. In this paper, we adopt the \emph{expected improvement} (EI) ~\cite{EI}, which determines new query points by maximising expected gain relative to the function values observed so far, although other options are possible ~\cite{review_bayes_opt}. 
At round $t$ of Algorithm~\ref{alg:BO}, EI is therefore given by $    \alpha_{\text{EI}}(\bm{x}|\mathcal{D}_t) = \mathbb{E}_{\text{GP-predictive}}[\max\{g(\bm{x})- g(\bm{x}_{t}^{+}), 0\}]$, where $\bm{x}_{t}^{+} = 
\arg\max_{\bm{x} \in \{\bm{x}_\ell\}_{\ell=1}^t}
g(\bm{x})$ and the expectation is computed using the posterior of the learnt GP (\ref{Eq:PredictiveDistribution}).
When $\bm{x}$ is continuous, the maximisation step in line 4 of Algorithm~\ref{alg:BO} can be executed using standard optimisation tools. 

\subsection{BOiLS: Bayesian Optimisation for Logic Synthesis}\label{Sec:BOILS}
The Bayesian optimisation machinery described in the previous section assumes continuously valued optimisation variables. Unfortunately, in logic synthesis, sequential and categorical optimisation variables render a direct deployment of BO inapplicable. Now, we introduce BOiLS, a logic-synthesis-specific BO algorithm that generalises recent works in combinatorial BO~\cite{casmo, BOSS} for sequential optimisation. BOiLS modifies GP kernels and acquisition maximisers to achieve state-of-the-art QoR results (see Section~\ref{sec:experiments}) as we detail next.
\subsubsection{GP Kernels for Logic Synthesis}\label{Sec:SSK} The first step during BO is building a GP surrogate model from QoR data $\mathcal{D}_t = \{\texttt{seq}_i, -\text{QoR}_{\mathcal{C}}(\texttt{seq}_i)\}_{i=1}^{n_t}$ with $n_t$ denoting the number of attempted sequences up-to round $t$. To do so, we assume that $-\text{QoR}_{\mathcal{C}}(\texttt{seq}) \sim \mathcal{G}\mathcal{P}(0, k_{\bm{\theta}}^{(\text{LS})}(\texttt{seq}, \texttt{seq}^{\prime}))$. Here, we used LS as the kernel's super-script to signify the need for new logic-synthesis functions that measure similarity between categorical sequences of operations applied to AIG graphs rather than between continuously-valued inputs. To define such kernels, we represent sequences in logic synthesis as strings of operations, with each character being an algorithm from \texttt{Alg}. Similar to~\cite{BOSS, SSK}, we measure the similarity between strings through the number of \emph{sub-strings} they have in common. Namely, we employ the sub-sequence string kernel (SSK) that uses sub-sequences of characters as similarity features. Formally, an $\ell^{th}$ order SSK between two strings $\texttt{seq}$ and $\texttt{seq}^{\prime}$ is defined as: $k_{\bm{\theta}}^{(\text{LS})}(\texttt{seq}, \texttt{seq}^{\prime}) = \sum_{\bm{u} \in \Sigma^{\ell}} c_{\bm{u}}(\texttt{seq})c_{\bm{u}}(\texttt{seq}^{\prime})$, where $\Sigma^{\ell}$ denotes the set of all possible ordered collections of up to $\ell$ characters from our alphabet. Moreover, $c_{\bm{u}}(\texttt{seq})$ measures the contribution of sub-sequence $\bm{u}$ to $\texttt{seq}$ which is defined using two tunebale hyperameters $\theta_{m} \in [0,1]$ and $\theta_{g}\in [0,1]$ that control the relative weighting of long and highly non-contiguous sub-strings:  
\begin{equation*}
    c_{\bm{u}}(\texttt{seq}) = \theta_{m}^{|\bm{u}|} \sum_{\substack{\bm{i} = (i_1, \dots, i_{|\bm{u}|})\\ 1 \leq i_1 < \dots < i_{|\bm{u}|} \leq |\texttt{seq}|}} \theta_{g}^{\operatorname{gap}(\bm{u}, \bm{i})}\mathbb{I}_{\bm{u}}(\texttt{seq}_{\bm{i}})
\end{equation*}
where $|\bm{u}|$ is the length of the sub-sequence, $\texttt{seq}_{\bm{i}} = (\texttt{seq}_{i_1}, \dots, \texttt{seq}_{i_{|\bm{u}|}})$,
$\operatorname{gap}(\bm{u}, \bm{i} ) = i_{|\bm{u}|}-i_1 + 1 - |\bm{u}|$, and $\mathbb{I}_{\bm{x}}(\bm{y})$ is the indicator function assessing if strings $\bm{x}$ and $\bm{y}$ match. We illustrate this kernel in Table~\ref{Table:Seq} on some logic synthesis sequences. For clarity, consider the first row and column in Table 1. First, we observe a match between $\bm{u}$ and $\texttt{seq}$. Given that $|\bm{u}| = 5$, we can already set $\theta_{m}^{5}$. Now, we notice that we can construct two matchings between $\bm{u}$ and $\texttt{seq}$ on indices $\bm{i} = (1, 2, 3, 6, 7)$ or $\bm{i}^\prime = (1, 2, 5, 6, 7)$. Therefore, the summation in the computation of $c_{\bm{u}}(\texttt{seq})$ runs over $\bm{i}$ and $\bm{i}^\prime$. In both cases, $\operatorname{gap}(u, \bm{i}) = \operatorname{gap}(\bm{u}, \bm{i}^\prime) = 2$ thus $c_{\bm{u}}(\texttt{seq}) = 2 \theta_{m}^{5}\theta_{g}^{2}$ in this case. Once  the kernel has been set, the  match and gap decays $\bm{\theta}= (\theta_m, \theta_g) \in [0,1]^2$ still have to be learnt from historical data $\mathcal{D}_{t} = \{\texttt{seq}_i, -\text{QoR}(\texttt{seq}_i)\}_{i=1}^{n_t}$. To do so, we make use of Equation~\ref{Eq:NLLM} while following projected gradients to ensure feasibility in the $[0,1]^2$ range: 
$\bm{\theta}_{\text{update}} = \text{Projection}_{[0,1]^{2}}\left(\bm{\theta}_{\text{current}} - \eta \nabla_{\bm{\theta}}\mathcal{J}(\bm{\theta}_{\text{current}})\right)$, with $\eta$ being a step-size. In practice, we implement the above update using a projected version of Adam~\cite{adam}. 

\begin{table}
{\scriptsize
\begin{tabular}{c|ccc}
                   & \multicolumn{3}{c}{Contribution of the sub-sequence ${\tt u}$}                                         \\
${\tt seq}$        & ${\tt RwRfDsBlRw}$                                 & ${\tt RwRfDsFr}$                & ${\tt RwRf}$                       \\ \hline
${\tt RwRfDsSoDsBl Rw}$ & 2$\theta_m^5\theta_g^2$                       & 0                             & $\theta_m^2$                      \\
                   & ${\tt \underline{RwRfDs} So Ds \underline{Bl Rw}}$ & -                             & ${\tt \underline{RwRf }Ds So Ds Bl Rw}$ \\ 
                   \cline{2-4} 
${\tt RwRfDsFrSoBlRw}$ & $\theta_m^5\theta_g^2$                       & $\theta_m^4$                 & $\theta_m^2$                      \\
                   & ${\tt \underline{RwRfDs}FrSo\underline{BlRw}}$      & ${\tt \underline{RwRfDsFr}SoBlRw}$ & ${\tt \underline{RwRf}DsFrSoBlRw}$      \\ \cline{2-4} 
${\tt RwRfDsFrBlSoBl}$    & 0                                              & $\theta_m^4$                 & $\theta_m^2$                      \\
                   & -                                              & ${\tt \underline{RwRfDsFr}BlSoBl}$ & ${\tt \underline{RwRf}DsFrFrSoBl}$      \\ \hline
\end{tabular}
}
\caption{Contribution $c_{\tt u}({\tt seq})$ of three sub-sequences in three sequences. ${\tt Rw, Rf, Bl, Fr, So, Bl, Ds}$ respectively stand for rewrite, refactor, balance, fraig, sopb, blut,  and dsdb.}
\label{Table:Seq}
\vspace{-3em}
\end{table}


\begin{algorithm}
\caption{BOiLS: BO for Logic Synthesis }
\begin{algorithmic}[1]
\State {\bf Input:} Circuit $\mathcal{C}$, maximum number of evaluations $N_{\max}$, maximum number of transformations per sequence $K$
\State {\bf Initialisation \& kernel tuning:}
\State Construct $\mathcal{D}_{0}=\{\texttt{seq}_i, \text{QoR}_{\mathcal{C}}(\texttt{seq}_i)\}_{i=1}^{N_{\text{init}}}$ by randomly sampling $N_{\text{init}}$ sequences
\State Set the TR radius to $R_{N_{\text{init}}}=K$
\State {\bf Optimisation loop:}
\For{$t=0, \dots, N_{\max}-1$}
\State Use $\mathcal{D}_{t}$ to fit a GP (Section~\ref{Sec:SSK})
\State Get ${\tt seq}_{t+1} \in \arg\max_{ {\tt seq} \in \text{TR}(\widehat{{\tt seq}}_t, \rho_t)}\alpha_{\text{EI}}({\tt seq}|\mathcal{D}_t)$ 
\State Evaluate $\text{QoR}_\mathcal{C}({\tt seq}_{t+1})$ \& augment data
\State Update the TRs radius $\rho_{t+1}$ (Section~\ref{Sec:TRS})
\EndFor
\State {\bf Ouptut:} The best sequence of operations  $\widehat{{\tt seq}}_{N_{\max}}$ found

\end{algorithmic}
\label{alg:BOiLS}
\end{algorithm}

\subsubsection{Trust-Region Local Search Acquisition Maximisers} \label{Sec:TRS}
As in standard BO, BOiLS executes an acquisition maximisation step after fitting a GP with the kernel above, effectively solving $\max_{\texttt{seq}\in \texttt{Alg}^{K}}\alpha_{\text{EI}}(\texttt{seq}|\mathcal{D}_t)$. The combinatorial nature of this acquisition maximisation step poses difficulties to global search techniques. To remedy those challenges, we equip BOiLS with a local search strategy around an adaptive trust region. At each round $t$, we use $\widehat{{\tt seq}}_t$ to denote the best sequence observed so far and define a trust-region as: 
$\text{TR}(\widehat{{\tt seq}}_t, \rho_t) = \{ \texttt{seq} \in {\tt Alg}^K : \text{Hamming}(\widehat{{\tt seq}}_t, \texttt{seq})  \leq \rho_t \}$,  where $\text{Hamming}(\texttt{a}, \texttt{b})$ is the Hamming distance counting the number of positions with different symbols between $\texttt{a}$ and $\texttt{b}$, and $\rho_t$ is an adjustable trust-region radius that we heuristically schedule as follows: 1) $\rho_{t} = \rho_{t-1} +1$  \text{if we observe 3 improving sequences in a row}, 2) $\rho_{t} = \rho_{t-1} - 1$ \text{if we observe 20 non-improving sequences in a row}, or 3) keep $\rho_{t}$ unchanged otherwise. In case, $\rho_t$ arrives at 0, the trust region is empty and the algorithm restarts in an attempt to avoid the current local minimum. With $ \text{TR}(\widehat{{\tt seq}}_t, \rho_t) $ defined, we use a simple local search strategy from~\cite{casmo} to maximise $\alpha_{\text{EI}}(\texttt{seq}|\mathcal{D}_t)$. Our strategy operates as follows: we randomly sample an initial configuration ${\tt seq}_0$ in the trust region and evaluate $\alpha_{\text{EI}}(\texttt{seq}_0|\mathcal{D}_t)$.
We then randomly select a neighbour point of a Hamming
distance 1 to ${\tt seq}_0$ in the TR, evaluate its acquisition function $\alpha_{\text{EI}}(\cdot|\mathcal{D}_t)$, and
move from ${\tt seq}_0$ if the neighbour has a higher acquisition function value. We repeat this process
until a preset budget of queries is exhausted and dispatch
the best configurations for objective function evaluation.


\section{Experimental Results}
\label{sec:experiments}
Now, we assess BOiLS' performance against existing automated and heuristic-based solutions on $10$ circuits from the set of EPFL arithmetic benchmarks \cite{epfl_benchmark}. Our results indicate that in 8 out of 10 circuits, BOiLS attains best QoR values across all methods while reducing sample complexities.
\begin{table*}[ht]
\begin{tabular}{l|cccccccc|cc}
 & \multicolumn{1}{l}{DRiLLS (PPO)} & \multicolumn{1}{l}{DRiLLS (A2C)} & \multicolumn{1}{l}{Graph-RL} & \multicolumn{1}{l}{GA} & RS & \multicolumn{1}{l}{Greedy} &  SBO & \multicolumn{1}{l|}{BOiLS} & \multicolumn{1}{l}{EPFL best (lvl)} & \multicolumn{1}{l}{EPFL best (count)} \\ \hline
Adder  &            22.62    & 24.59            & 24.48             & 24.80         & 24.27     & 23.36     & 25.02          & \textbf{25.57}    & 21.36 & -55.76  \\
Barrel Shifter    & \textbf{00.00}    & \textbf{00.00}            &  \textbf{00.00}             &  \textbf{00.00}         &  \textbf{00.00}     &  \textbf{00.00}     &  \textbf{00.00}          & \textbf{00.00}    &  \textbf{00.00} &  \textbf{00.00}   \\
Divisor           & 40.40    & 42.66            & -                 & 44.82         & 43.78     & 40.46     & 45.49          & \textbf{47.36}    & -59.52          & 14.04   \\
Hypotenuse        & 00.81    & 00.89            & -                 & 01.66         & 01.75     & -0.04     & 01.77          &\textbf{5.99}      & -68.80          & 01.62   \\
Log2              & 07.02    & 07.48            & -                 & 07.96         & 07.77     & 04.70     & \textbf{09.01} & 08.70             & 06.25 & -33.34  \\
Max               & 29.28    & 30.49            & 31.51             & 31.97         & 30.76     & 28.14     & 31.04          & 31.77             & \textbf{35.61}  & -164.0  \\
Multiplier        & 18.56    & 19.15            & -                 & 20.20         & 19.25     & 18.32     & 20.33          & \textbf{21.13}    & 20.67 & 00.00   \\
Sine              & 01.64    & 02.18            & 01.64             & 02.70         & 01.88     & 00.79     & 02.64          & \textbf{03.82}    & -23426.71       & -26.21  \\
Square-root       & 12.47    & 14.07            & 13.23  & 13.06         & 13.70     & 08.19     & 13.79          & \textbf{14.10}    & 00.00 & 11.14   \\
Square            & 36.65    & 37.77            & 37.88  & 38.01         & 37.78     & 36.56     & 38.27          & \textbf{38.90}    & 38.88  & -21.81  \\ \hline
Average           & 16.94    & 17.93            & -   & 18.52         & 18.09     & 16.05     & 18.77          & \textbf{19.74}    & -2343 & -29.66                     
\end{tabular}
\end{table*}

\begin{figure*}[ht]
\centering
\includegraphics[trim = {0em 0em 0em 0.5em}, clip=true, width=.9\textwidth]{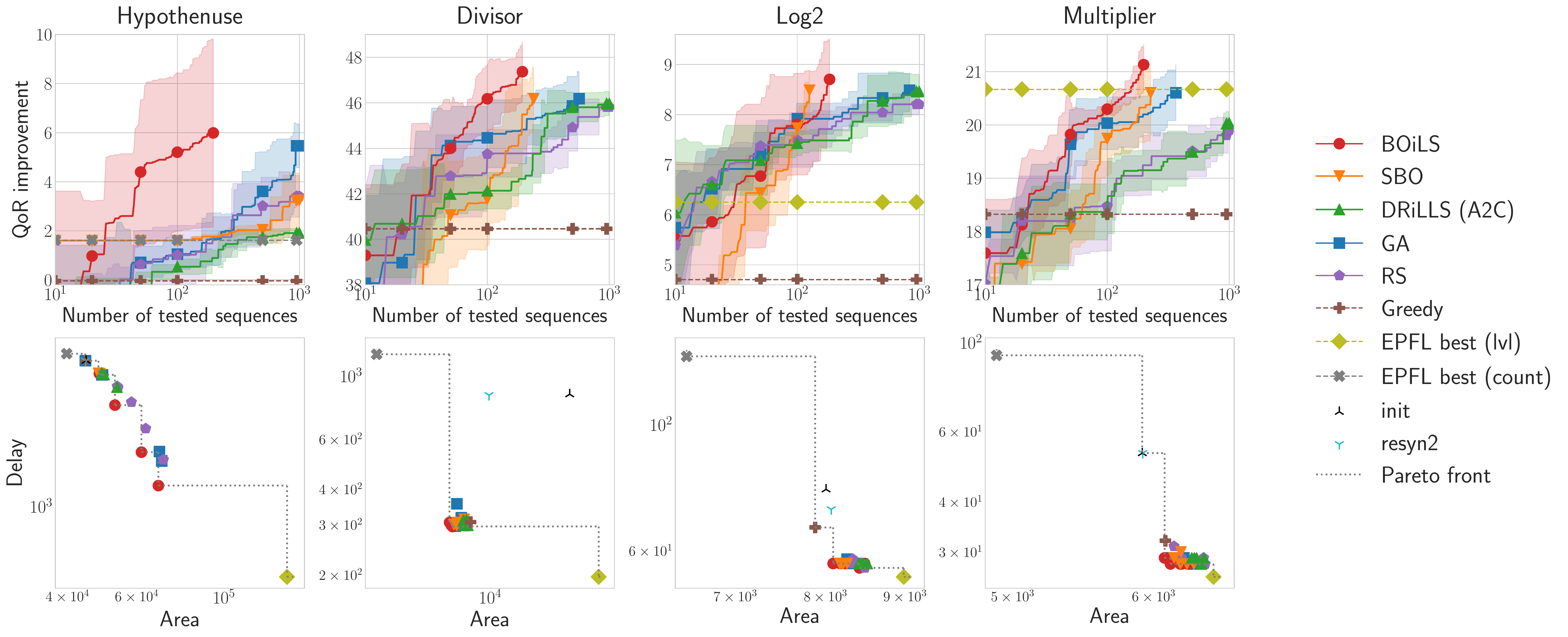}
\caption{(Top Row): Tabular report of QoR improvement (in \%) for all ten circuits averaged over five random seeds. Please note that high computational demands associated with extracting large circuit graphs limit the applicability of graph RL algorithms to small settings. (Middle Row): Results on the four largest circuits demonstrating that BOiLS acquires improved QoR results after about 200 iterations. Standard BO and GAs present competitive baselines. DRL, on the other hand, rarely outperforms RS strategies. (Bottom Row): Pareto front comparisons on the same four large circuits with a restricted budget of Nmax=200 evaluations.}
\end{figure*}




\subsection{Experimental setup}
Our experiments were performed on two machines with Intel Xeon CPU E5-2699 v4@2.20GHz, 64GB RAM, running Ubuntu 18.04.4 LTS and equipped with one NVIDIA Tesla V100 GPU. 
All algorithms were implemented in Python 3.7 relying on \texttt{ABC v1.01}. Area and delay characteristics were measured after FGPA mapping (performed through \texttt{if -K 6} command) using the \texttt{print\_stats}  command of \texttt{ABC}. For each circuit $\mathcal{C}$, we ran BOiLS and alternative synthesis flow tuning methods to solve the optimisation problem in Equation~(\ref{eq:best_solution}) with $K=20$ primitive transformations that included the following algorithms: $\texttt{Alg}$ = [{\tt  rewrite}, {\tt rewrite -z}, {\tt refactor}, {\tt refactor -z}, {\tt resub}, {\tt resub -z}, {\tt balance}, {\tt fraig}, {\tt sopb}, {\tt blut}, {\tt dsdb}]. 
We compared BOiLS to a large set of solvers and ran each experiment across five random seeds to record statistically significant results for all the optimisers we considered:

\begin{itemize}[leftmargin=0.15in, itemsep=0.05pt, topsep=0.05pt]
    \item \textbf{Deep Reinforcement Learning}: We benchmarked against DRiLLS~\cite{drills} and Graph-RL~\cite{GCN_for_eda}. In terms of Graph-RL, we followed the work in~\cite{GCN_for_eda}, and for DRiLLS we employed the code provided by the authors~~\cite{drills} attempting both PPO and A2C policy update rules. We modified the rewards to account for our goal from Equation~(\ref{eq:best_solution}). 
    \item \textbf{Standard Bayesian optimisation (SBO):} To assess the importance of designing logic-synthesis specific kernels and acquisitions introduced by BOiLS, we also included standard BO as a benchmark relying on the implementation from \cite{HEBO}.
    \item \textbf{Genetic Algorithm (GA):} 
    Rather than building surrogate models, one could also solve (\ref{eq:best_solution}) using genetic algorithms that support mutation and cross-over. Although such methods are known to be more sample intensive, we included GA algorithms from \cite{geneticalgo2} as additional baselines to understand how sequences from BOiLS compare to those generated by evolutionary search.
    \item \textbf{Random Search (RS):} Although generally omitted, we add random search as a baseline. Our implementation relied on the Latin hypercube samplers from \texttt{pymoo} \cite{pymoo}. 
    \item \textbf{Greedy Algorithm:} We also contrast with a greedy algorithm, which 
    builds a unique sequence of length $K$ by appending transformations that provide the largest immediate QoR improvement.
    \item \textbf{EPFL best (count / lvl)}: Those results are the  best known solutions achieved for each circuit in~\cite{epfl_benchmark}. Of course, current heuristics disjointly consider area (\textit{count}) or delay (\textit{lvl}), where no one heuristic can simultaneously optimise both. As such, those aggregated values form a new baseline. 
\end{itemize}

\subsection{Experimental Results}
Next, we provide answers to the following three questions: 
\begin{itemize}[leftmargin=0.15in, itemsep=0.05pt, topsep=0.05pt]
\item \textbf{Q.I.} With budget constraints, does BOiLS produce new state-of-the-art QoRs? 
\item \textbf{Q.II.} If given a higher budget, would other methods improve? 
\item \textbf{Q.III.} Do generated sequences belong to the Pareto-Front between area and delay?
\end{itemize}
\subsubsection{BOiLS is Efficient with High QoRs} In this section, we affirmatively answer \emph{both} $\textbf{Q.I}$ and $\textbf{Q.II.}$ 

\underline{\textbf{A.I. State-of-the-Art QoRs:}} The table in the top row of Figure 3 reports the best-achieved QoR results across all circuits while restricting the interaction budget $N_{\text{max}} = 200$ across all algorithms. Those values are averaged over five random seeds and computed as a relative improvement (in \%) compared to resync2 using $ (
\text{QoR}_\mathcal{C}({\tt resyn2}) - \text{QoR}_\mathcal{C}(\widehat{{\tt seq}}_t) )/\text{QoR}_\mathcal{C}({\tt resyn2})$. From this table, we can see that BOiLS achieves the best results on average over 8/10 designs and that SBO is best in $\texttt{Log2}$ circuits and is mostly second to BOiLS. Such results indicate that BO is a vital alternative to consider in logic-synthesis and that the sequential modifications from Section~\ref{Sec:BOILS} further improve performance.
Finally, we remark that DRL-based approaches and greedy strategy perform comparably to RS. 

\underline{\textbf{A.II. Sample Efficiency:}} 
We ran additional experiments to assess sample efficiency, increasing the allowable budget for all other algorithms except for BOiLs. The goal was to understand how many trials would take algorithms to recover the QoRs offered by BOiLS. We terminated the loop if methods achieved 97.5\% values of BOiLS QoRs or until exhausting a total of $N_{\max} = 1000$ iterations. We report those results in the middle row of Figure 3 on five large circuits; average results on all 10 has been previously shown in Figure 1. We realise that: 1) SBO recovers our QoRs but requires $1.5$ more trials, 2) GA algorithms need $2.8$ times more attempts than BOiLS, and 3) DRL necessitate over $5$ times additional sample complexity. 

\underline{\textbf{A Remark on RS as a Valuable Baseline:}} Our results show that RS is a competitive baseline. We noticed that RS provides similar results to DRL even after 1000 trials. We also ran GA for an $N_{\max}$ of 1000 to assess further improvements. We realised that after 1000 trials, GA attains 4.3\% improvement to RS while being $\approx 1\%$ worst than BOiLS. DRL on the other hand, only achieved $\approx 0.12\%$ improvements to RS. We urge the community to consider RS as an alternative baseline when operating ML techniques.

\subsubsection{BOiLS solutions are Pareto-Efficient} 
Finally, we investigated the area and delay 
profiles provided by each algorithm over each circuit. The bottom row of Figure 3 displays the profiles obtained by the best found solutions after $N_{\max}= 200$ iterations for each of the five seeds. Considering 5 random seeds in those large circuits, we show that solutions from BOiLS are on the Pareto front 55\% of the time, compared to 20\% for SBO, 15\% in GA, and 0\% for RS and DRL.



\section{Conclusion \& Future Work}
We proposed BOiLS, the first modern Bayesian optimisation solver for logic synthesis applications. BOiLS utilises sequential kernels and trust-region constrained acquisition optimisers to search in combinatorial spaces. Our empirical results signify the importance of BO methodologies in logic synthesis, demonstrating improved QoR values and reduced sample complexities. Although we chose to optimise QoRs, we note that BOiLS is not tied to a specific black-box and can be utilised with other quantities of interest, e.g., area or delay disjointly by simply modifying Equation~(\ref{Eq:QoR}). 

In future work, we plan to extend BO beyond logic synthesis to other steps in the electronic design automation workflow. 


\begin{thebibliography}{00}


\bibitem{QoR_max}
E. Testa \textit{et al.},
''
Extending Boolean Methods for Scalable Logic Synthesis
,''
IEEE Access, 2020.

\bibitem{review_logic_synthesis}
E. Testa \textit{et al.},
''
Logic synthesis for established and emerging computing
,''
Proceedings of the IEEE, 2018.

\bibitem{pimap}
G. Liu \textit{et al.},
''
PIMap: A Flexible Framework for Improving LUT-Based Technology Mapping via Parallelized Iterative Optimization
,''
ACM Transactions on Reconfigurable Technology and Systems (TRETS), 2019.

\bibitem{himap}
D. Wijerathne \textit{et al.},
''
HiMap: Fast and Scalable High-Quality Mapping on CGRA via Hierarchical Abstraction
,'' in \textit{DATE}, 2021.

\bibitem{ml_design_reduction}
S. Ellouz \textit{et al.},
''
Combining internal probing with artificial neural networks for optimal RFIC testing
,''
IEEE International Test Conference, 2006.


\bibitem{ml_placement}
S. Ward \textit{et al.},
''
PADE: A high-performance placer with automatic datapath extraction and evaluation through high-dimensional data learning
,''
DAC Design Automation Conference, 2012.

\bibitem{ml_routing}
Z. Xie \textit{et al.},
''
RouteNet: Routability prediction for mixed-size designs using convolutional neural network
,''
IEEE/ACM International Conference on Computer-Aided Design (ICCAD), 2018.

\bibitem{ml_testing}
S. Fine \textit{et al.},
''
Coverage directed test generation for functional verification using bayesian networks
,''
Proceedings of the 40th annual Design Automation Conference, 2003.

\bibitem{ml_verification}
H-G. Stratigopoulos \textit{et al.},
''
Error moderation in low-cost machine-learning-based
analog/RF testing
,''
IEEE Transactions on Computer-Aided Design of Integrated Circuits and Systems, 2008.

\bibitem{survey_ml_for_eda} G. Huang 
\textit{et al.},
''
Machine learning for electronic design automation: A survey
,''
ACM Transactions on Design Automation of Electronic Systems (TODAES).


\bibitem{cnn_for_eda}
C. Yu \textit{et al.},
''
Developing Synthesis Flows Without Human Knowledge
,''
Proceedings of the 55th Annual Design Automation Conference, 2018.




\bibitem{drills} H. Abdelrahman, S. Hashemi \textit{et al.}
''
DRiLLS: Deep reinforcement learning for logic synthesis
,''
2020 25th Asia and South Pacific Design Automation Conference (ASP-DAC).


\bibitem{GCN_for_eda}
W. Haaswijk \textit{et al.},
''
Deep Learning for Logic Optimisation Algorithms
,''
IEEE International Symposium on Circuits and Systems (ISCAS), 2018.

\bibitem{model_based_atari}
L. Kaiser \textit{et al.},
''
Model-based reinforcement learning for atari
,''
ICLR, 2020.


\bibitem{epfl_benchmark}
L. Amaru, P.-E. Gaillardon, and G. De Micheli,
''
The epfl combinational ´
benchmark suite
,''
IWLS, no. CONF, 2015.




\bibitem{casmo} X. Wan \textit{et al.},
''
Think Global and Act Local: Bayesian Optimisation over High-Dimensional Categorical and Mixed Search Spaces
,''
ICML 2021.

\bibitem{yosys} 
C. Wolf and J. Glaser,
''
Yosys - A Free Verilog Synthesis Suite
''
in Proceedings of Austrochip 2013.

\bibitem{yu2020decision}
C. Yu and W. Zhou,
''
Decision Making in Synthesis cross Technologies using LSTMs and Transfer Learning
,''
Proceedings of the 2020 ACM/IEEE Workshop on Machine Learning for CAD.


\bibitem{ABC}
A. Mishchenko \textit{et al.},
''
Abc: A system for sequential synthesis and
verification
,''
URL http://www. eecs. berkeley. edu/alanmi/abc, pp. 1–
17, 2007.

\bibitem{LazyMan}
W. Yang \textit{et al.},
''
Lazy man's logic synthesis
,''
IEEE/ACM International Conference on Computer-Aided Design (ICCAD), 2012.

\bibitem{pymoo}
J. Blank and K. Deb,
''
Pymoo: Multi-Objective Optimisation in Python
,''
IEEE Access, 2020.

\bibitem{geneticalgo2}
D. Pascal,
''geneticalgorithm2 (v.6.2.12)'',
https://github.com/PasaOpasen/geneticalgorithm2, 2021.

\bibitem{submodular_opt}
A. Krause \textit{et al.},
''
Submodular function maximization.
,''
Tractability, 2014.

\bibitem{review_bayes_opt}
B. Shahriari \textit{et al.},
''
Taking the human out of the loop: A review of Bayesian optimization
,''
Proceedings of the IEEE, 2015.

\bibitem{HEBO}
A. I. Cowen-Rivers \textit{et al.}, 
''An Empirical Study of Assumptions in Bayesian Optimisation,''
arXiv preprint arXiv:2012.03826, 2020.

\bibitem{James}
J. Hensman \textit{et al.}, 
''Gaussian processes for big data,''
Twenty-Ninth Conference on Uncertainty in Artificial Intelligence (UAI2013)


\bibitem{Rasmussen}
C. E. Rasmussen \textit{et al.},
''
Gaussian Processes for Machine Learning (Adaptive Computation and Machine Learning)
,''
The MIT Press, 2005.

\bibitem{EI}
J. Moˇckus,
''
On Bayesian methods for seeking the extremum
,'' in Optimization Techniques IFIP Technical Conference, pp. 400–404, Springer, 1975.

\bibitem{BOSS}
H. B. Moss \textit{et al.},
''
BOSS: Bayesian Optimization over String Spaces
,''
in Advances in Neural Information Processing Systems (NeurIPS), 2020.

\bibitem{adam}
D. Kingma \textit{et al.},
''
Adam: A method for stochastic optimization
,''
ICLR, 2015.

\bibitem{SSK}
H. Lodhi \textit{et al.},
''
Text classification using string kernels
,''
Journal of Machine Learning Research, 2002.

\end{thebibliography}
\end{document}